\documentclass[sigconf]{acmart}

\usepackage{amsmath}
\usepackage{bbold}
\usepackage{mathtools}
\usepackage{colortbl} 
\usepackage{arydshln} 
\usepackage{balance}
\usepackage{color}
\usepackage{bbding}
\usepackage{url}
\usepackage{multirow}
\usepackage{algorithmicx,algorithm}
\usepackage{algorithm}
\usepackage{algpseudocode}
\usepackage{amsmath}
\AtBeginDocument{%
  \providecommand\BibTeX{{%
    \normalfont B\kern-0.5em{\scshape i\kern-0.25em b}\kern-0.8em\TeX}}}

\copyrightyear{2022}
\acmYear{2022}
\setcopyright{acmcopyright}
\acmConference[MM '22] {Proceedings of the 30th ACM International Conference on Multimedia}{October 10--14, 2022}{Lisboa, Portugal.}
\acmBooktitle{Proceedings of the 30th ACM International Conference on Multimedia (MM '22), Oct. 10--14, 2022, Lisboa, Portugal}
\acmPrice{15.00}
\acmISBN{978-1-4503-9203-7/22/10}
\acmDOI{10.1145/3503161.3548054}

\acmSubmissionID{1278}

\begin{document}
\renewcommand{\algorithmicrequire}{\textbf{Input:}}
\renewcommand{\algorithmicensure}{\textbf{Output:}} 

\title{Class Gradient Projection For Continual Learning}


\author{Cheng Chen}
\affiliation{%
  \institution{Center for Future Media, University of Electronic Science and Technology of China}
  \city{Chengdu}
  \country{China}
}
\email{cczacks@gmail.com}

\author{Ji Zhang}
\affiliation{%
  \institution{Center for Future Media, University of Electronic Science and Technology of China}
  \city{Chengdu}
  \country{China}
}
\email{jizhang.jim@gmail.com}

\author{Jingkuan Song\textsuperscript{\rm 1,2}}
\authornote{Corresponding Author.}
\affiliation{%
  \institution{\textsuperscript{1}Shenzhen Institute for Advanced Study, University of Electronic Science and Technology of China \\
  \textsuperscript{2}Peng Cheng Laboratory}
  \city{Shenzhen}
  \country{China}
}
\email{jingkuan.song@gmail.com}

\author{Lianli Gao}
\affiliation{%
  \institution{Shenzhen Institute for Advanced Study, University of Electronic Science and Technology of China}
  \city{Shenzhen}
  \country{China}
}
\email{lianli.gao@uestc.edu.cn}
\renewcommand{\shortauthors}{Cheng Chen, Ji Zhang, Jingkuan Song, \& Lianli Gao}
\begin{abstract}
  Catastrophic forgetting is one of the most critical challenges in Continual Learning (CL). 
  Recent approaches tackle this problem by projecting the gradient update orthogonal to the gradient subspace of existing tasks. 
  While the results are remarkable, those approaches ignore the fact that these calculated gradients are not guaranteed to be orthogonal to the gradient subspace of each class due to the class deviation in tasks, 
  e.g., distinguishing “Man” from “Sea” v.s.  differentiating “Boy” from “Girl”. 
  Therefore, this strategy may still cause catastrophic forgetting for some classes.
  In this paper, we propose Class Gradient Projection (CGP), which calculates the gradient subspace from individual classes rather than tasks.  
  Gradient update orthogonal to the gradient subspace of existing classes can be effectively utilized to minimize interference from other classes. 
  To improve the generalization and efficiency, we further design a Base Refining (BR) algorithm to combine similar classes and refine class bases dynamically.
  Moreover, we leverage a contrastive learning method to improve the model's ability to handle unseen tasks. 
  Extensive experiments on benchmark datasets demonstrate the effectiveness of our proposed approach. It improves the previous methods by 2.0\% on the CIFAR-100 dataset. 
  The code is available at \url{https://github.com/zackschen/CGP}.
\end{abstract}




\maketitle

\section{Introduction}
\label{intro}

\begin{figure}[t]
	\centering
	\includegraphics[width=\linewidth]{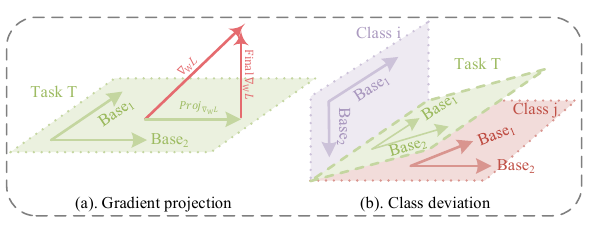}
	\caption{(a) In the standard gradient projection methods for CL, 
	the actual parameters update is $Final\nabla_WL$, which ensures that the network does not forget the knowledge learned from the previous task.
	(b) The class deviation can simultaneously affect the bases calculation and the gradient projection, as a consequence, causing catastrophic forgetting.}
	\Description{The learn update projection.}
	\label{projection}
\end{figure}
Although the modern deep learning network (DNN) has shown good performance on the task at hand \cite{DBLP:conf/mm/CaiDZWWWL20,DBLP:conf/mm/WangH0YHWMW20,zhang2022progressive,DBLP:conf/mm/DengL20,DBLP:conf/mm/LiLT0YW0H20,DBLP:conf/mm/ZhangSYG21}, they always violently forget the previous knowledge when they learn new tasks. Such a phenomenon is known as \textit{catastrophic forgetting}, which is the research hotspot in Continual Learning (CL) \cite{mccloskey1989catastrophic,ratcliff1990connectionist,delange2021continual}.
In the training phase of CL, the model needs to learn on a stream of tasks sequentially.
This requires the network to have \textit{plasticity}, that is, the ability to learn fresh knowledge from unseen tasks. 
At the same time, it is critical for the network to retain previously learned knowledge, \textit{i.e.}, the \textit{stability} ability.
However, “you can't have your cake and eat it too”: the dilemma of stability-plasticity is what the continual learning networks need to solve primarily.

Many approaches have been proposed to deal with this dilemma, including i) regularization-based methods \cite{kirkpatrick2017overcoming,aljundi2018memory}, ii) memory-based methods \cite{lopez2017gradient,chaudhry2018efficient}, and iii) expansion-based methods \cite{abati2020conditional,li2019learn}.
To be more specific, regularization-based methods focus on penalizing the modification of the most important weights of the network so that the previous knowledge can be preserved.
To maintain the performance on previous tasks, memory-based approaches replay the data of old tasks or the synthetic data from generative models to jointly train with present task samples. 
In contrast, expansion-based methods seek to expand the architecture of the network to learn new knowledge.
\begin{figure*}[t]
	\centering
	\includegraphics[width=0.95\linewidth]{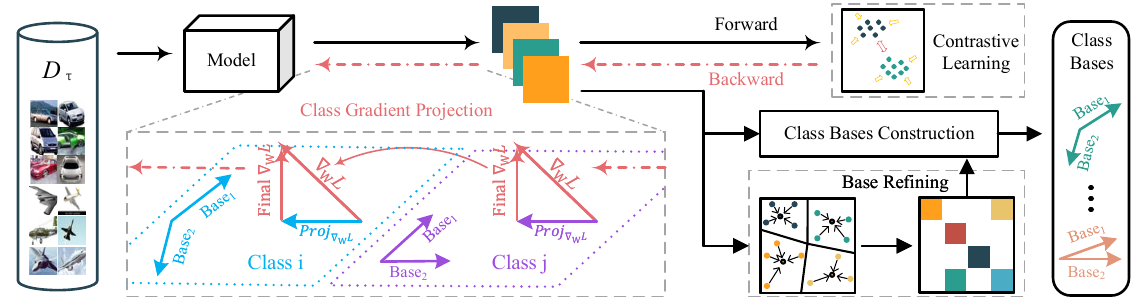}
	\caption{An overview of our \textbf{Class Gradient Projection (CGP)} network. 
	First, to alleviate the class deviation, CGP constructs the gradient subspace for each class with a base refining module.
	Second, CGP projects the gradient of the new task orthogonal to the constructed gradient subspace.
	Moreover, we develop a contrastive loss to guide continuous optimization of the model when its gradient is projected by more and more constructed gradient subspace.}
	\Description{Framework}
	\label{framework}
\end{figure*}

Recently, the gradient projection based paradigms \cite{farajtabar2020orthogonal,zeng2019continual,saha2021gradient} have achieved remarkable performance in tackling the catastrophic forgetting problem in CL.
Existing gradient projection based methods put explicit constraints on the gradient directions that the optimizer takes. 
For example, GPM \cite{saha2021gradient} computes the bases of these gradient subspace based on the representations learned by each learning task. 
In the learning phase of the next task, new gradient steps in the orthogonal direction to these gradient subspace would be taken.
A simple illustration is shown in Fig.~\ref{projection} (a).
Despite the impressive progress, there is a critical problem for those gradient projection based methods.
Considering a simple scenario where the task at hand is to classify the given image as “Man” or “Sea”. 
The class deviation between them is significantly large.
In other words, their gradient subspace could be so different from each other. 
Therefore, if we directly perform Singular Value Decomposition (SVD) \cite{lee2018self} on the representations from all classes of a task, some bases for “apple” or “car” may be missed.
When the gradient projection is performed in the subsequent tasks, the projected gradient is greatly disturbed; 
as a consequence, the model forgets the knowledge learned previously, as illustrated in Fig.~\ref{intro} (b).

In this paper, we propose Class Gradient Projection (CGP) to address the side-effect of class deviation in gradient projection.
CGP calculates the gradient subspace with the representations from individual classes rather than tasks, such that the gradient update steps orthogonal to the constructed classes subspace can be effectively utilized to minimize the interference from other classes.
Based on this framework, two effective variants are developed to further alleviate catastrophic forgetting.
Specifically, to construct a more informative gradient subspace for each incoming task, we design a Base Refining (BR) algorithm to refine class bases dynamically. 
BR adopts an inter-class similarity estimation module to guide the partition of classes into the built bases. 
In addition, considering that the update steps are orthogonal to the gradient subspace of previous classes, the model may lack optimization space for unseen classes.  
Therefore, we introduce a supervised contrastive loss to learn more representative and robust features by pulling embedding from positive samples closer and pushing embedding from negative samples apart. 
The contrastive learning can guide continuous optimization of the model when its gradient is projected by more and more classes subspace.
The designed two variants can complement each other to enhance the plasticity and stability of the model simultaneously.
In summary, the main contributions can be summarized as follows:
\begin{itemize}
\item We propose Class Gradient Projection (CGP), which effectively utilizes the gradient update steps orthogonal to the constructed class subspace to minimize the negative interference between classes.

\item Based on the CGP framework, we introduce a Base Refining (BR) algorithm as well as a contrastive loss  to further alleviate catastrophic forgetting. The two components can complement each other to enhance the plasticity and stability of the model simultaneously.

\item We conduct extensive experiments on several benchmark datasets and various network architectures.  
The achieved results demonstrate the effectiveness of our method. 
\end{itemize}

\section{Related Work}

\subsection{Continual Learning}
Prevalent approaches to address the catastrophic forgetting problem in continual learning can be grouped into three broad categories: regularization-based approaches \cite{aljundi2018memory,zenke2017continual,nguyen2017variational,lee2017overcoming}, 
expansion-based approaches \cite{rusu2016progressive,veniat2020efficient,hung2019compacting,li2019learn,rajasegaran2019random,serra2018overcoming}, and memory-based approaches \cite{ayub2021eec,lopez2017gradient,sprechmann2018memory,rebuffi2017icarl}.

Regularization-based methods aim at avoiding excessive changes in the parameters learned on old tasks when learning a new task.
Typically, these methods \cite{aljundi2018memory,zenke2017continual} estimate importance weights for each model parameter. 
Then the changes of the important parameters are penalized by a regularizer for previous tasks.
Under the Bayesian framework, VCL and IMM \cite{nguyen2017variational,lee2017overcoming}  take the posterior distribution of network parameters learned from previous tasks as the prior distribution of network parameters on the current task, which implicitly penalizes the changes of network parameters under the Bayesian framework.
However, our method works in putting constraints on gradient descent rather than ascribing importance to parameters and penalizing the changes.

The basic idea of expansion-based approaches is to directly add or modify the model structure. Some methods (e.g. \cite{rusu2016progressive}) add a network to each task and lateral connections to the network of the previous task. MNTDP \cite{veniat2020efficient} proposes a modular layer network approach, whose modules represent atomic skills that can be composed to perform a certain task and provides a learning algorithm to search the modules to combine with. 
These strategies may work well, but they are computationally expensive and memory intensive.
Finally, other approaches \cite{rajasegaran2019random,serra2018overcoming} provide an alternative approach that assigns the different sub-networks or weights.
In contrast to these methods, our method works within a fixed network architecture.

For memory-based methods, catastrophic forgetting is avoided by storing data from previous tasks and training them together with data from the current task. 
Some methods \cite{lopez2017gradient,sprechmann2018memory,rebuffi2017icarl} use replayed samples from previous tasks to constrain the parameters' update when learning the new task. 
They do not need to store data from previous tasks, but their performance is significantly affected by the quality of generated data, especially for complex natural images.
Instead of saving the raw data or generated data, our method saves the subspace bases for future task learning, avoiding the privacy problem.

Recently, a series of continual learning methods combined with orthogonal projection have been proposed \cite{farajtabar2020orthogonal,saha2021gradient}. 
Gradient projection methods update the model with gradients in the orthogonal directions of old tasks, without access to old task data.
GPM \cite{saha2021gradient} find the bases of these subspaces by analyzing network representations after learning each task with Singular Value Decomposition (SVD) in a single shot manner and store them in the memory as gradient projection memory.
In this paper, to address the class deviation in gradient projection, we propose class gradient projection.
Through such class gradient projection, the network achieves better stability and preserve more learned knowledge.

\subsection{Contrastive Learning}
In recent years, contrastive learning has shown superior performance, even in competition with supervised training. Supervised contrastive learning (SCL) is extended from the standard contrastive loss by incorporating the label information to construct positive and negative pairs \cite{DBLP:journals/corr/abs-2001-07685}. 
Prototypical Contrastive Learning (PCL) \cite{li2020prototypical} uses the centroids of clusters as prototypes, and pulls the image embedding closer to its prototypes. 
Noise-Contrastive estimation \cite{gutmann2010noise} is the seminal work that estimates the latent distribution by contrasting it with artificial noises.
In addition, CPC \cite{van2018representation} tries to learn representations from visual inputs by leveraging an auto-regressive model to predict the future in an unsupervised manner.
These studies \cite{chen2020simple,chen2021exploring} have resolved practical limitations that have previously made learning difficult such as negative sample pairs, large batch size, and momentum encoders.
Meanwhile, it has been shown that supervised learning can also enjoy the benefits of contrastive representation learning by simply using labels to extend the definition of positive samples \cite{khosla2020supervised}.
Another method \cite{Cha_2021_ICCV} combines the contrastive learning with continual learning.
They use samples of the current task as anchors and samples of previous tasks as negative samples.
Different from this method, our method does not need to replay the previous samples.
Instead, we perform augmentation on input samples.
The sample and the augmented sample make up the positive pairs.
The rest samples serve as negative samples. 
Then we introduce a contrastive loss to pull embedding from positive samples closer and to push embedding from negative samples apart. 
This contrastive learning encourages the network to learn more representative and robust features of tasks.

\section{Preliminaries}

\subsection{Continual Learning}
In the setup of supervised continual learning, a series of $T$ tasks are learned sequentially. 
We denote the task by its task descriptor, $\tau \in \{1,2,...,T\}$ and its corresponding dataset $\textit{D}_\tau = \{{(x_{\tau,i},\mathit{y}_{\tau,i} )}_{i=1}^{N_\tau}\}$ which has $n_\tau$ example pairs.
The $x_{\tau,i}$ ($\in \textit{X}$ ) is the input vector and $\mathit{y}_{\tau,i}$ ($\in \textit{Y}$ ) is the target vector. 
A DNN model parameterized with $\Phi = \{\mathbb{W},\varphi\}$ is used to learn a mapping in the form $\hat{\mathit{y}_\tau} = \mathit{f_{map}}(x_\tau;\Phi)$. Here, $\mathbb{W}_\tau$ = $\{\mathit{W^l_\tau}\}_{l=1}^L$ represents a $L$ layer neural network, where $\mathit{W}^l_\tau$ is the layer-wise weight for layer $\mathit{l}$ and task $\tau$. 
In each layer, the layer network computes the output $\mathit{x_{\tau,i}^{l+1}}$ for next layer:
\begin{displaymath}
\mathit{O_{\tau,i}^l} = \mathit{f}(x^l_{\tau,i};\mathit{W_\tau^l}), \quad \mathit{x_{\tau,i}^{l+1}} = \sigma_l(\mathit{O_{\tau,i}^l}),
\end{displaymath}
with $\sigma_l$ is a non-linear function for layer $l$ and $\mathit{f}$ is the linear function for layer. Following \cite{saha2021gradient}, at the first layer, the $\mathit{x_{\tau,i}^{1}}$ = $\mathit{x_{\tau,i}}$ represents the raw input data from task $\tau$.
Whereas in the
subsequent layers we define $\mathit{x_{\tau,i}^l}$ as the representations of $\mathit{x_{\tau,i}}$ at layer $l$.
The output of final neural network $\mathit{x_{\tau,i}^{L+1}}$ is then passed through a classifier parameterized by $\varphi$ to produce the prediction $\hat{\mathit{y}} = \mathit{f}(\mathit{x_{\tau,i}^{L+1}};\varphi)$.
The model is trained by minimizing the loss function for task $\tau$, e.g. cross-entropy loss
\begin{displaymath}
 \Phi^* = \underset{\Phi}{\textit{minimize}}\sum_{n=1}^{N_t}\ell ( f(\mathit{x_{\tau,i}},\Phi),\mathit{y_{\tau,i}} ),
\end{displaymath}
where $\Phi^*$ denotes the optimal model for task $\tau$.

\subsection{Gradient Projection}
Recently, a series of continual learning methods combined with orthogonal gradient projection have been proposed \cite{farajtabar2020orthogonal,saha2021gradient}. These methods update the model with gradients in the orthogonal directions of old tasks, without access to old task data.
After learning the task $\tau$ completely, they construct the gradient space $\mathit{S_\tau}$ using the samples of task $\tau$. 
When learning the task $\tau+1$, the gradient of model $\mathbb{W}_{\tau+1}$ is projected to the gradient subspace $\mathit{S}_\tau$ of previous tasks to get the $\textit{Proj}_{\nabla_{\mathbb{W}_{\tau+1}L}}$.
Then the $\textit{Proj}_{\nabla_{\mathbb{W}_{\tau+1}L}}$ is substracted out from the origin gradient $\mathbb{W}_{t+1}$ so that remaining gradient updates lie in the space orthogonal to $S_\tau$  \cite{saha2021gradient}.


\section{Method}
\label{method}

As illustrated in the preliminaries, the gradient projection approaches construct the gradient subspace by the samples from all the classes of task $\tau$. However, we argue that class deviation in tasks may cause the calculated gradient space too biased to represent the task accurately, leading to the degradation of the performance. 
Moreover, it is critical to explore the optimization space for the new tasks.
In this section, we propose a novel Class Gradient Projection (CGP) for continual learning.
Fig.~\ref{framework} illustrates the pipeline of our proposed continual learning approach.
We are going to show how the CGP works at a high level.
On the one hand, CGP constructs the gradient subspace with individual classes and projects the gradient update of new tasks to the direction orthogonal to the subspace of old classes.
On another hand, CGP learns the representations with supervised contrastive learning to explore the optimization space for the new tasks.
The training is done on the compound loss:

\begin{equation}
	\label{loss}
	\mathcal{L}  = \mathcal{L}_{ce} + \lambda \cdot \mathcal{L}_{con} ,
\end{equation}
where $\mathcal{L}_{ce}$ is the cross-entropy loss and $\mathcal{L}_{con}$ is the supervised contrastive learning.

\subsection{Class Gradient Projection}

In this section, we introduce the bases construction and class gradient projection to show how it enables the network to learn continually without forgetting.

\paragraph{When learning Task {$\tau$ = 1}:} 
We use the training process as shown in Eq.~\ref{loss} without any constraint. 
After the network learns completely, the Singular Value Decomposition (SVD) is performed on the representations to construct the bases. 
Specifically, for layer $l$ in $\mathbb{W}$, we construct a representation matrix $\mathit{R_1^l} = [x_{1,1}^l,x_{1,2}^l,...,x_{1,r}^l]$ from the samples of a certain number $r$.
We separate the matrix $\mathit{R_1^l}$ by the target label:
\begin{displaymath}
	\mathit{R_1^l} = [C_{1,1}^l,C_{1,2}^l,...,C_{1,c}^l,...,C_{1,c_\tau}^l],
\end{displaymath}
 with
\begin{displaymath}
	C_{1,c}^l = [x_{1,c,1}^l,x_{1,c,2}^l,...,x_{1,c,n_{c}}^l],
\end{displaymath}
where $c_\tau$ represents the class number in task $\tau$ and $n_{c_\tau}$ is the number of samples in class $c_\tau$.
Next, we perform SVD on $C_{1,1}^l \in \mathbb{R}^{m \times n}$:
\begin{equation}
\label{SVD}
\mathit{U}_{1,1}^l,\Sigma_{1,1}^l,(\mathit{V}_{1,1}^l)^T = SVD(C_{1,1}^l),
\end{equation}
where $\mathit{U}_{1,1}^l$ is a $m \times m$ complex unitary matrix, and $\mathit{V}_{1,1}^l$ is a $n \times n$ complex unitary matrix. $\Sigma_{1,1}^l$ is a $m \times n$ rectangular diagonal matrix with non-negative singular values $\{\delta\}$ on the diagonal in a descending order.
By applying the k-rank approximation on $C_{1,1}^l$ for the given threshold ( $\epsilon^l_\tau$ ):
\begin{equation}
	\label{k-rank1}
	\left \| (\mathit{C_{1,1}^l})_k \right \| _F^2 \ge \epsilon_\tau ^l\left \| \mathit{C_{1,1}^l} \right \| _F^2,
\end{equation}
where $\left \| \cdot \right \|_F$  is the Frobenius norm, we construct the bases $S_1^l$ by picking the first $k$ vectors in $\mathit{U}_{1,1}^l$. 

When learning the rest classes, e.g. class $j$, we need to remove the common bases contained in $S_1^l$ so that the bases constructed by $C_{1,j}^l$ are orthogonal to the existing bases in $S_1^l$:
\begin{equation}
\label{projectbase}
\hat{C}_{1,j}^l = C_{1,j}^l - S_1^l ({S_1^l})^T (C_{1,j}^l) = C_{1,j}^l - \textit{Proj}_{C_{1,j}^l}.
\end{equation}
After applying the SVD on $\hat{C}_{1,j}^l$, we follow the criteria to construct the bases for class $j$:
\begin{equation}
	\label{k-rank2}
	\left \| \textit{Proj}_{C_{1,j}^l} \right \| _F^2 + \left \| (\mathit{\hat{C}_{1,j}^l})_k \right \| _F^2 \ge \epsilon_\tau ^l\left \| \mathit{\hat{C}_{1,j}^l} \right \| _F^2.
\end{equation}
After calculating all classes for task 1, we construct the bases $S_1^l$.

\paragraph{When learning Task {$\tau$ > 1}:}
We train the network using the loss function defined in Eq.~\ref{loss} as usual. 
However, before taking gradient updates from the backpropagation, we put constraints on the gradient updates. We modify the origin gradients using:
\begin{equation}
\label{project1}
\textit{Proj}_{\nabla_{\mathbb{W}}\mathcal{L}} = ({\nabla_{\mathbb{W}}\mathcal{L}}) S^l ({S^l})^T.
\end{equation}
Hence, the final gradient can be expressed as:
\begin{equation}
\label{project2}
\nabla_{\mathbb{W}}\mathcal{L} = {\nabla_{\mathbb{W}}\mathcal{L}} - \textit{Proj}_{\nabla_{\mathbb{W}}\mathcal{L}}.
\end{equation}
After the network training converged, we construct the bases for task $\tau$. 
We use the same process as the calculation in task $\tau = 1$, shown in Eq.~\ref{projectbase} and Eq.~\ref{k-rank2}.
After the construction, we concatenate the bases for task $\tau$, i.e. $S_\tau^l$ and previous bases $S^l$ together in a new $S^l$ which used for future learning.
We repeat bases construction and gradient projection until all tasks are learned.

\subsection{CGP with Base Refining (BR)}
\label{method_simi}
There are many classes in a task $\tau$, as illustrated in Sec.~\ref{intro}.
Some classes are different from others.
However, there may be some classes similar to others too. 
For example, the man is different from the sea and similar to the boy. 
So we introduce the base refining to combine similar classes to construct the bases of gradient subspace.
We estimate the class similarity using the prototype of class. 
Concretely, after the task $\tau$ is learned, we collect the samples $\mathit{X}_{c_\tau}$ of all classes.
Then the class prototype is calculated by the normalized mean embedding:
\begin{equation}
\label{prototype}
	\mathit{z}^l_{c_\tau} = \frac{1}{\left |\mathit{X}_{c_\tau} \right | } \sum_{i \in \mathit{X}_{c_\tau}}x_{i}^l , \quad
	\mathit{P}^l_{c_\tau} = \frac{\mathit{z}^l_{c_\tau}}{\left \| \mathit{z}^l_{c_\tau} \right \|_2 } ,
\end{equation}
where $x_{i}^l$ is the representation of sample at layer $l$.
The similarity between classes is estimated by the calculated prototype using cosine distance:
\begin{equation}
\label{similarityEstimation}
	sim (\mathit{P}^l_{u},\mathit{P}^l_{v}) = \frac{{\mathit{P}^l_{u}}^T\mathit{P}^l_{v}}{\left \| \mathit{P}^l_{u} \right \|\left \| \mathit{P}^l_{v} \right \|  } .
\end{equation}
If the $sim (\mathit{P}^l_{u},\mathit{P}^l_{v})$ is greater than threshold $\eta$, we construct the bases for both class $u$ and $v$.
In contrast, if the similarity is smaller than $\eta$, we construct the bases for class $u$ and $v$ as before.
We introduce how to combine the base refining with our base construction in the next paragraph.


When constructing bases in $\tau=1$, as we can get the representations of classes, the representations of class $j$ and $k$ are concatenated together, $C_{1,(j,k)}^l = concat(C_{1,j}^l,C_{1,k}^l)$. 
Then, we perform the SVD on $C_{1,(j,k)}^l$.
Following the criteria in Eq.~\ref{k-rank1}, we construct the bases $S_{1,(j,k)}^l$ for both class $j$ and class $k$.
Finally, we replace the origin bases $S_{1,j}^l$ and $S_{1,k}^l$ with $S_{1,(j,k)}^l$.
In the construction for task $\tau>1$, we construct the common bases for class $j$ and $k$ as in \cite{DBLP:journals/corr/abs-2202-02931}.
Firstly, we calculate the square of the singular value of $C_{\tau,j}^l$ with respect to $S^l$ which are constructed by previous class by: 
\begin{equation}
\label{commonbase}
	\delta_{\tau,j}^l = S^l C_{\tau,j}^l (C_{\tau,j}^l)^T (S^l)^T.
\end{equation}
Then, the SVD is applied to the result of performing Eq.~\ref{projectbase} on $C_{\tau,j}^l$, $\hat{C}_{\tau,j}^l = \mathit{\hat{U}}_{\tau,j}^l\hat{\Sigma}_{\tau,j}^l(\mathit{\hat{V}}_{\tau,j}^l)^T$, where $\hat{\Sigma}_{\tau,j}^l$ contains singular values $\{ \hat{\delta}_{\tau,j}^l \}$.
Next, we concatenate the $\{\delta_{\tau,j}^l\}$ and $\{ (\hat{\delta}_{\tau,j}^l)^2 \}$ together in a vector $\delta$.
By performing k-rank ( Eq.~\ref{k-rank2} ) on $C_{\tau,j}^l$, we choose the corresponding bases of first $k$ elements in $\delta$ to be the bases for class $j$ and $k$.

Furthermore, in practice, one does not need all $\mathit{X}_{c_\tau}$ for calculation. 
Another alternative is to select the samples $\mathit{X}_{right_{c_\tau}}$ which are predicting right to the ground truth label.
This alternative is referred to as BR-GTL (Ground Truth Label).
Furthermore, BR-GTL reduces the storage size and calculate consumption.
We empirically observe that BR-GTL slightly outperforms the BR-STD (Standard) which uses samples no matter the prediction result.
The BR-GTL is used in all of our following experiments.

\subsection{CGP with Contrastive Learning (Con)}
Although class gradient projection can reserve the knowledge well, it reduces the optimization space for learning fresh knowledge.
To deal with this problem, we introduce the contrastive learning to explore the optimization space for new tasks.
Specifically, given the $N_\tau$ samples from task $\tau$,
we apply augmentation to each sample and obtain 2$N_\tau$ inputs $\{x_{\tau,i}\}_{i=1}^{2N_\tau}$.
The augmentation consists of color and brightness changes with details given in Sec.~\ref{setting}.
We collect the normalized embeddings $\{ \mathit{x_{\tau,i}^{L+1}}\}_{i=1}^{2N_\tau}$ before the classifier.
The contrastive learning loss is defined as:
\begin{equation}
	\mathcal{L}_{con} =  \sum_{i=1}^{N_\tau } -log\frac{exp(x_i \cdot{} x_{j(i)} / \mu  )}{ {\textstyle \sum_{k=1}^{2N_\tau}}\mathbb{1}_{i\ne k} exp(x_i \cdot{} x_k / \mu  )  } ,
\end{equation}
where $x_{j(i)}$ is the augmentation input from the same source image $x_i$, and $\mu$ is the scalar temperature parameter.
Through this contrastive learning, we can pull the embeddings between the pair of positive samples $x_i$ and $x_{j(i)}$ closer, while pushing the embeddings with the $2(N_\tau - 1)$ pairs of negative samples apart.
This contrastive learning encourages the network to learn a discriminative representation that is robust to low-level image corruption.

\begin{algorithm}[t]
\caption{Algorithm for Class Gradient Projection.}
\label{algorithm1}
\begin{algorithmic}[1]
    \Require
          Model $\Phi = \{\mathbb{W},\varphi\}$;  Task sequence $T$; Class base memory $S$; Class prototype memory $P$;  Learning rate $\alpha$; Similarity threshold $\eta$; Threshold $\epsilon$; Sample size for Base Refining $n_{r}$; Train epoch $E$. 
    \Ensure
      Model $\Phi = \{\mathbb{W},\varphi\}$; Class base memory $S$; Class prototype memory $P$.
    \State Initialize Model $\Phi$: $\Phi \leftarrow \Phi_0$.
    \For {$\tau=0,\ldots,\left | T \right | $}
    	\For {$e_\tau=0,\ldots,E$} \algorithmiccomment{train and project} 
        	\State $B_n \sim D_\tau$  \algorithmiccomment{sample a mini-batch of size n from task $\tau$.} 
        	\State $B_{2n} \sim AUGMENT(B_n)$ 
        	\algorithmiccomment{augment samples}
        	\State	$\nabla_{\mathbb{W}}\mathcal{L} \leftarrow OPTIMIZER(B_{2n},\Phi)$
     	    \State $\nabla_{\mathbb{W}}\mathcal{L} \leftarrow  PROJECT(\nabla_{\mathbb{W}}\mathcal{L},S)$ \algorithmiccomment{Eq.~ (\ref{project1},\ref{project2})}
     	    \State $\Phi \leftarrow \Phi - \alpha \nabla_{\mathbb{W}}\mathcal{L}$
    	\EndFor
    	\For{$c=0,\ldots,c_\tau$}{} 
    	\State $B_c \sim D_\tau$ \algorithmiccomment{sample prediction right of size $n_{r}$ for class $c$.} 
    	\State ${C_c} \leftarrow FORWARD(B_c,\Phi)$ \algorithmiccomment{construct representation}
    	\State $\mathit{P}_c \leftarrow CALCULATE(C_c)$ \algorithmiccomment{Eq.~\ref{prototype}}
    	\State $simi(\mathit{P}_c,P) \leftarrow CALCULATE(\mathit{P}_c, P)$ \algorithmiccomment{Eq.~ \ref{similarityEstimation}}
    	\State $\hat{C_c} \leftarrow PROJECT(C_c,S)$ \algorithmiccomment{Eq.~\ref{projectbase}}
    	\State $U_c,\Sigma_{c}\mathit{V}_{c}^T \leftarrow SVD(\hat{C_c})$ \algorithmiccomment{Eq.~\ref{SVD}}
    	\If {$Simi(\mathit{P}_c,P) < \eta$}
    	\State $k \leftarrow CRITERIA(\hat{C_c},C_c,\epsilon_\tau)$ \algorithmiccomment{Eq.~ (\ref{k-rank1},\ref{k-rank2})}
    	\State $S \leftarrow [S,U_c[0:k]]$
    	\Else
        \State $\delta_{c} \leftarrow CALCULAT(C_c,S)$ \algorithmiccomment{Eq.~\ref{commonbase}}
        \State $k \leftarrow CRITERIA(C_c,\epsilon_\tau)$ \algorithmiccomment{Eq.~\ref{k-rank1}}
        \State $\delta \leftarrow SORT([\delta_{c},\Sigma_{c}^2])$
        \State $S \leftarrow [S,[U_c,S](\delta[0:k])]$ \algorithmiccomment{select first k bases corresponding to the first k elements in $\delta$}
    	\EndIf
    	\State $P \leftarrow [P,\mathit{P}_c]$
    	\EndFor
    	
    \EndFor
    \State \Return $\Phi,S,P$
\end{algorithmic}
\end{algorithm}

Algorithm \ref{algorithm1} describes the details of our CGP combining with the base refining and contrastive learning.

\section{Experiments}
\label{setting}
To evaluate the effectiveness of our proposed method, we first compare it with state-of-the-art CL methods. Then, we conduct ablation studies to empirically analyze the main components.
\subsection{Experimental setting}
Let us first describe the experimental settings on datasets, comparison methods, evaluation metrics and implementation details.
\subsubsection{Datasets}
Following \cite{saha2021gradient}, we evaluate on several continual learning benchmarks, including \textbf{10-Split CIFAR-100} \cite{krizhevsky2009learning}, \textbf{20-Split CIFAR-100} \cite{krizhevsky2009learning}, \textbf{5-Split CIFAR-100} \cite{krizhevsky2009learning}, \textbf{CIFAR-100 Superclass}, and sequence of \textbf{5-Datasets} \cite{ebrahimi2020adversarial}.
The 10-Split CIFAR-100 is constructed by randomly splitting 100 classes of CIFAR-100 into 10 tasks with 10 classes per task.
The 20-Split CIFAR-100 consists of splitting CIFAR-100 into 20 tasks, each one with 5 classes.
The 5-Split CIFAR-100 is constructed by splitting the CIFAR-100 into 5 tasks, each task consisting of 20 classes.
The 10-Split CIFAR-100 is constructed by randomly splitting the 100 classes of CIFAR-100 into 10 tasks with 10 classes per task.
CIFAR-100 Superclass \cite{yoon2019scalable} split the 100 classes of CIFAR-100 into 20 tasks where each task has 5 different but semantically related classes.
Moreover, we also use the 5-Datasets which consists of CIFAR-10, MNIST, SVHN \cite{netzer2011reading}, notMNIST \cite{bulatov2011notmnist} and Fashion MNIST \cite{xiao2017fashion}. Each dataset in 5-Datasets is regarded as a learning task.

\subsubsection{Comparison methods} 
We compare our method with various continual learning methods including memory-based approaches and regularization-based approaches. 
Concretely, the memory-based approaches include reservoir sampling (ER\_Res) \cite{DBLP:journals/corr/abs-1902-10486},  Averaged GEM (A-GEM) \cite{chaudhry2018efficient}, and Orthogonal Weight Modulation (OWM) \cite{zeng2019continual}. For regularization-based methods, we use HAT \cite{serra2018overcoming} and Elastic Weight Consolidation (EWC) \cite{kirkpatrick2017overcoming}.
Besides, the state-of-the-art gradient projection method Gradient Projection Memory (GPM) \cite{saha2021gradient} is also adopted for comparison.
Additionally, we add the “Multitask” baseline where all tasks are learned jointly using the entire dataset at once in a single network. Multitask serves as the upper bound on average accuracy on all tasks.

\subsubsection{Evaluation metrics}
Following \cite{saha2021gradient}, we evaluate the performance on the following metrics: Average Accuracy (ACC) and Backward Transfer (BWT). ACC is the average test classification accuracy of all tasks. BWT measures the model’s capability of retaining previous
knowledge after learning a new task. Formally, ACC and BWT are defined as:
\begin{equation}
	ACC=\frac{1}{T} \sum_{i=1}^TA_{T,i}, \quad   BWT=\frac{1}{T-1} \sum_{i=1}^{T-1}A_{T,i}-A_{i,i}
\end{equation}
where T is the total number of sequential tasks, $A_{T,i}$ is the evaluated accuracy of the model on $\tau=i$ task after learning the $\tau=T$ task sequentially.

\subsubsection{Implementation details}
Following the general experiment setting of CL \cite{saha2021gradient,yoon2019scalable}, in our experiments, we use a 5-layer AlexNet for the 5-Split, 10-Split and 20-Split CIFAR-100 dataset.
For CIFAR-100 Superclass, we use the LeNet-5 architecture.
As for the 5-datasets, similar to \cite{DBLP:journals/corr/abs-1902-10486}, we use a reduced ResNet18 architecture. 
In the -Split CIFAR-100 and 5-Datasets experiments, we train each task for a maximum of 200 and 100 epochs respectively. 
The early termination strategy is also adopted as in \cite{serra2018overcoming}.
For all the datasets, we set the batch size 64.
We set the values of $\lambda$ and $\eta$ to 0.1 and 0.7, respectively.
In the network training stage, all tasks share the same backbone network but each task has its own classifier. The classifier is fixed after the model is trained on the corresponding task.
At inference, the task identifier can not be accessed.
We use the threshold $\epsilon$ in \cite{saha2021gradient} for SVD k-rank approximation. 
For the augmentation scheme in contrastive learning, we use AugMix \cite{hendrycks2020augmix} as the augmentation.

\subsection{Experimental Results}
Here, we present the quantitative evaluation results on various benchmark datasets and network structures to investigate our method.

\label{result}
\begin{table}[]
	\renewcommand\arraystretch{1.3}
	\renewcommand\tabcolsep{4.0pt}
	\caption{Performance comparison of CL methods in terms of ACC (\%) and BWT (\%) on CIFAR-100 and 5-Datasets.}
	\label{tab:result}
\begin{tabular}{ccccc}
	\hline
	\multirow{2}{*}{\textbf{Methods}} &  \multicolumn{2}{c}{\textbf{CIFAR-100}} & \multicolumn{2}{c}{\textbf{5-Datasets}} \\ 
	\cline{2-5} 
	 & ACC & BWT & ACC & BWT \\ \hline 
	OWM\cite{zeng2019continual} & 50.94 & -0.30 & - & - \\
	HAT\cite{serra2018overcoming} & 72.06 & -0.00 & \textbf{91.32} & -0.01 \\
	A-GEM\cite{chaudhry2018efficient} & 63.98 & -0.15 & 84.04 & -0.12 \\
	ER\_Res\cite{DBLP:journals/corr/abs-1902-10486} & 71.73 & -0.06 & 88.31 & -0.04 \\
	EWC\cite{kirkpatrick2017overcoming} & 68.80 & -0.02 & 88.64 & -0.04 \\
	GPM\cite{saha2021gradient} & 72.25 & 0.17 & 90.44 & -1.41 \\
	\textbf{CGP (ours)}  & \textbf{74.26} & -0.37 & 90.94 & -1.48 \\ \hline
	Multitask & 79.58 & - & 91.54 & - \\ \hline
\end{tabular}
\end{table}

\subsubsection{Comparison Results on CIFAR-100 and 5-Datasets}
Quantitative comparisons with state-of-the-art methods on
CFAR-100 and 5-Datasets are shown in Tab.~\ref{tab:result}. 
As can be observed from the table, on the CIFAR-100 dataset, our CGP consistently outperforms all baselines.
It is worth noting that our method produces around 2.0\% gain in terms of ACC compared to the outstanding competitor GPM. 
As discussed in Sec.~\ref{method}, the main difference between CGP and GPM is the way to calculate the bases. The comparison results demonstrate that there is severe class deviation in tasks, in addition, our method can solve this problem effectively.
Furthermore, the results achieved by CGP corroborates with our motivation that it is more beneficial to consider the class deviation in the task.
We also observe that even on the more challenging 5-Datasets, our method achieves the accuracy of 90.94\%, which is 0.5\% higher than that of the strong baseline GPM.

\begin{table}[]
	\renewcommand\arraystretch{1.3}
	\renewcommand\tabcolsep{4.0pt}
	\caption{Experiments result on CIFAR-100 Superclass dataset. $(\dagger)$ denotes the result reported from APD \cite{yoon2019scalable}. }
	\label{tab:superclass}
\begin{tabular}{ccc}
	\hline
	\textbf{Methods} & ACC (\%) & Capacity (\%) \\ \hline
	PGN$(\dagger)$ & 50.76 & 271 \\
	DEN$(\dagger)$ & 51.10 & 191 \\
	RCL$(\dagger)$ & 51.99 & 184 \\
	APD$(\dagger)$ & 56.81 & 130 \\
	\textbf{CGP(ours)} & \textbf{57.53} & \textbf{100} \\ \hline
	Multitask$(\dagger)$ & 61.00 & 100 \\ \hline
\end{tabular}
\end{table}

\subsubsection{Comparison Results on CIFAR-100 Superclass}
To compare our method with the state-of-the-art expansion based methods, we conduct experiment on CIFAR-100 Superclass dataset.
In this dataset, we split the CIFAR-100 to make each task contains 5 different but semantically related classes.
Comparison results are shown in Tab.~\ref{tab:superclass}, where “Capacity” denotes the percentage of network capacity used with respect to the original network.
As shown, our CGP outperforms other methods with a fixed capacity network, which suggests that gradient projection with class bases is indeed helpful for improving both plasticity and stability.
Specifically, our method achieves 57.53\% and -0.26\% in terms of ACC and BWT with the smallest network, respectively.
For instance, CGP outperforms APD with around 0.72\% gain in terms of ACC by 30\% fewer network parameters, showing the good implementability of our method.
To sum up, our approach successfully preserves the knowledge learned before and learns the useful representations for future learning, and thus it significantly mitigates catastrophic forgetting.

\subsection{Ablation Studies}
To analyze our method in more depth, we study the impact of different ablation variants of CGP on performance.
\label{ablationstudy}
\subsubsection{Components Analysis.}

\begin{table}[]
	\renewcommand\arraystretch{1.3}
	\renewcommand\tabcolsep{4.0pt}
	\caption{Ablation study on the designed components. “BR” and “Con” represents the base refining and contrastive learning, respectively.}
	\label{tab:ablation}
\begin{tabular}{ccccc}
\hline
CGP & BR &  Con & ACC(\%) & BWT(\%) \\ \hline
  &  & & 58.32 & -22.96 \\
 \Checkmark  & &  & 69.19 & \textbf{-0.08} \\
 \Checkmark &  \Checkmark & \multicolumn{1}{l}{} & 72.61 & -1.64 \\
 \Checkmark & \Checkmark & \Checkmark & \textbf{74.26} & -0.37 \\ \hline
\end{tabular}
\end{table}

\begin{figure*}[h]
	\centering
	\includegraphics[scale=0.4]{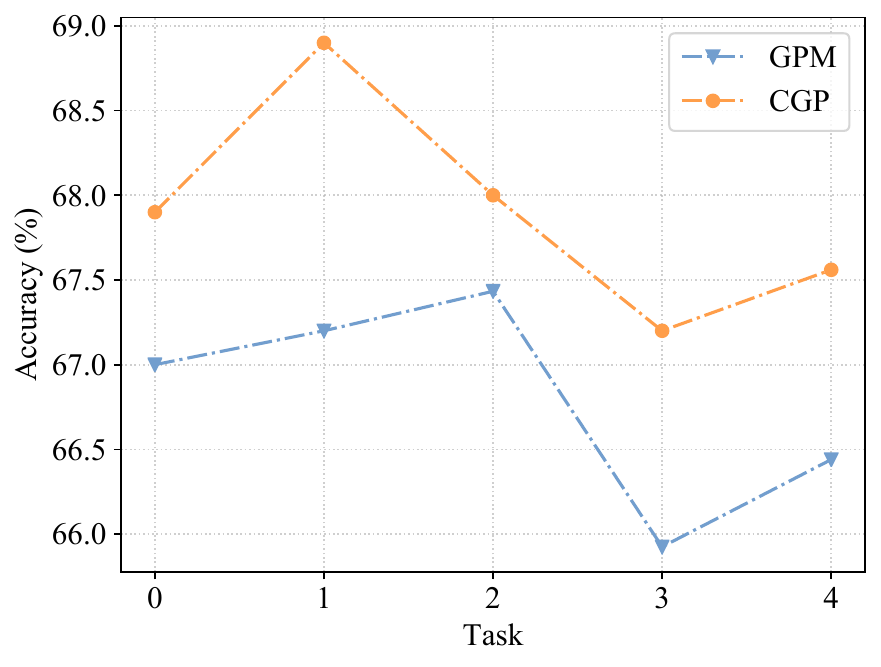}
	\includegraphics[scale=0.4]{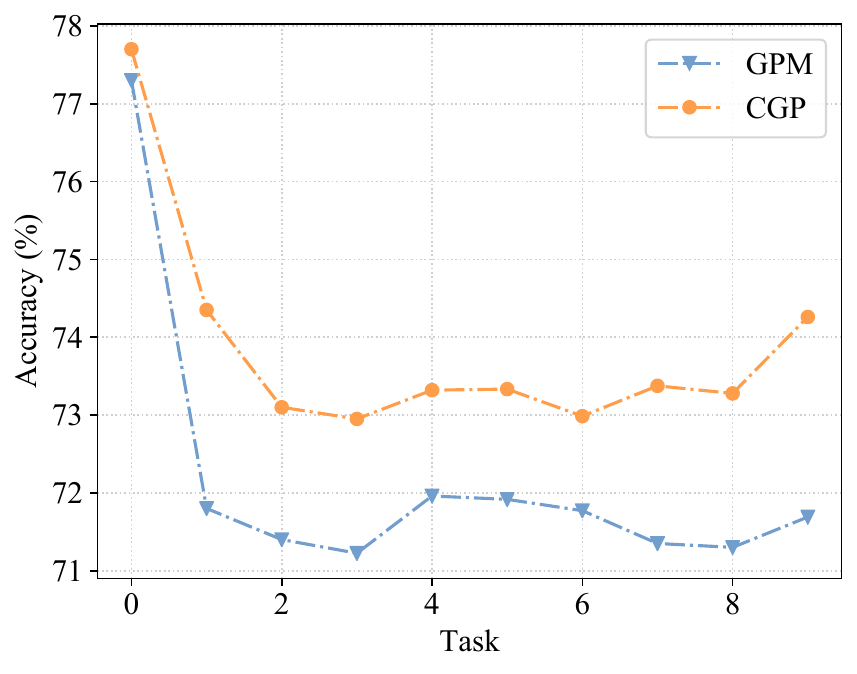}
	\includegraphics[scale=0.4]{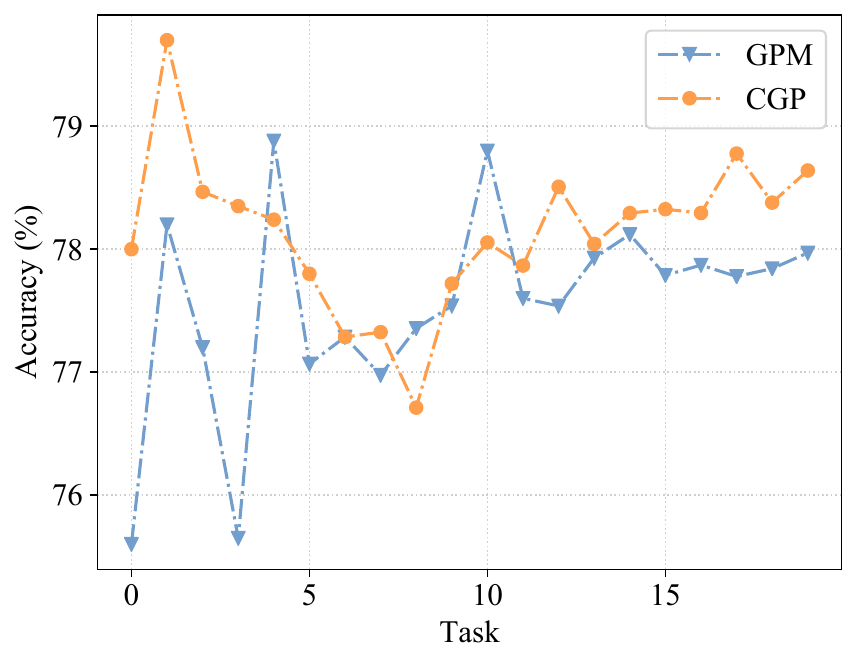}
	\caption{Performance comparison between GPM and CGP in terms of ACC (\%) on CIFAR-100 dataset with 5-Split (left), 10-Split (center) and 20-Split (right). We report the test results after these methods learn one task completely.}
	\Description{Framework}
	\label{sequenceresult}
\end{figure*}

\begin{figure}[h]
	\centering
	\includegraphics[width=0.9\linewidth]{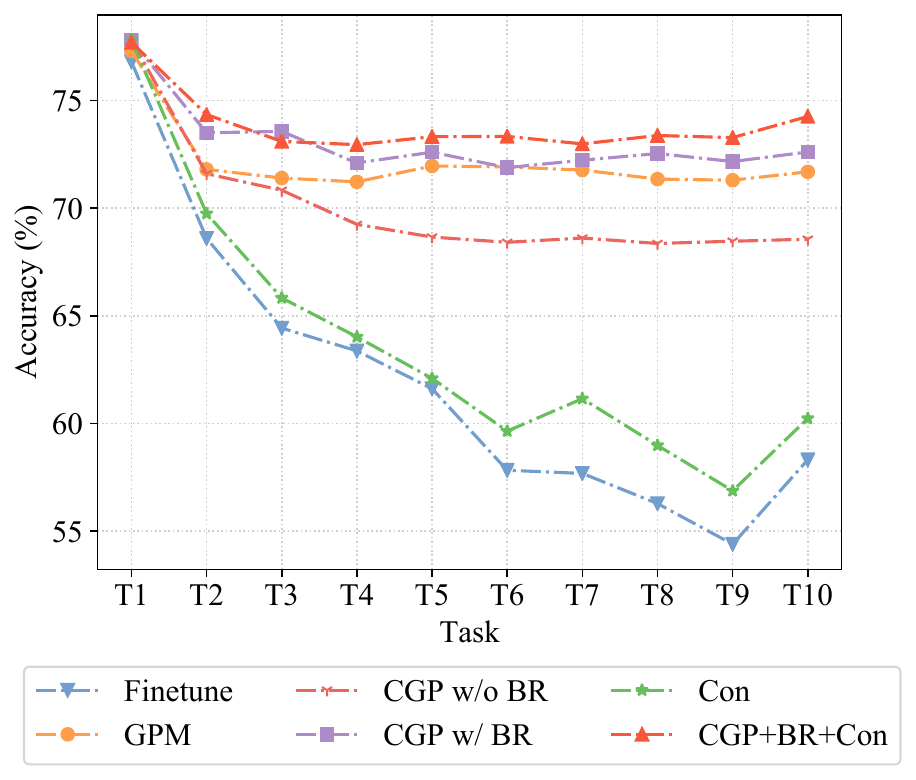}
	\caption{Performance comparison in terms of ACC (\%) on 10-Split CIFAR-100 dataset with different variants of our method.  } 
	\Description{Framework}
	\label{ablationfigure}
\end{figure}

We perform an ablation experiment on the 10-Split CIFAR-100 dataset to scrutinize  the effectiveness of different versions of CGP.
The ablative results are described in Tab.~\ref{tab:ablation}.
Note that the first row in the table represents the result of finetuning, which trains a single model to solve all the tasks without adding any component.
It can be observed from the results that finetuning suffers from catastrophic forgetting 
(drops to -22.96\% in terms of BWT).
Oppositely, we can observe that Class Gradient Projection brings a significant performance gain (row 2), producing around 10.87\% and 22.86\% increase in terms of ACC and BWT, respectively.
In particular, the our CGP achieves the best results on BWT (-0.08\%), suggesting its capability of preserving the knowledge of the learned tasks.
When CGP equipped Base Refining (row 3), we can observe our method achieves a better result of 72.61\% in terms of ACC. 
This comparison indicates that estimating the similarity between classes has an positive impact on the class gradient projection and maintaining a Base Refining structure of bases calculation is of particular importance.
In addition, when we combine contrastive learning with our CGP, the approach achieves the best performance, demonstrating the ability of our approach to learn new knowledge and preserve old knowledge.
Furthermore, Fig.~\ref{ablationfigure} indicates a more detailed results in terms of the average accuracy of the components after each task is completed.
We plot the learning curve of GPM to compare with our methods.
As shown in the figure, without contrastive learning, our CGP is still better than GPM.
Furthermore, when CGP is equipped with contrastive learning, it outperforms the GPM by a sizable margin.
This comparison indicates that our method can ensure less forgetting of old knowledge when learning knowledge from new tasks.

\begin{table}[]
	\renewcommand\arraystretch{1.3}
	\renewcommand\tabcolsep{4.0pt}
	\caption{Ablation study on the similarity threshold. We conduct experiments on the CIFAR-100 dataset.}
	\label{tab:threshold}
\begin{tabular}{ccccccc}
\hline
Threshold & 0.5 & 0.6 & 0.7 & 0.8 & 0.9 & 1.0 \\ \hline
ACC(\%) & 74.12 & 73.98 & \textbf{74.26} & 73.43 & 70.76 & 69.19 \\
BWT(\%) & -1.78 & -1.36 & -0.37 & -1.22 & -2.30 & \textbf{-0.08} \\ \hline
\end{tabular}
\end{table}

\subsubsection{Analysis on Stability and Plasticity.}
To study the balance of stability and plasticity, which is controlled by similarity threshold $\eta$, we compare the performance of our CGP by varying $\eta = $ 0.6, 0.7, 0.8, 0.9, 1.0 on 10-Split CIFAR-100 dataset.
As shown in Tab.~\ref{tab:threshold}, 
when the $\eta =$ 1.0, the backward transfer (BWT) becomes the best, just -0.08\%, which means that the network hardly forgets knowledge.
But the average accuracy comes to 69.19\%, this result clearly indicates the method just preserves the knowledge learned before, refusing to accept new knowledge.
When the $\eta$ decreases to 0.9, although the ACC is better, the BWT is worse. 
Maybe the relaxation of constraints increases the optimization space, disturbing the learning of new knowledge and the preserving of old knowledge.
With the decrease of $\eta$, the balance is going to be better.
So the performance is going to be better no matter ACC and BWT.
When the $\eta$ equals 0.7, the performance comes to the best with 74.26\% in ACC and -0.37\% in BWT.
Since ACC is affected by both stability and plasticity, showing that the network can learn more new knowledge and preserve the learned knowledge at the same time.
In addition, as the $\eta$ decreases, the BWT is going to be worse, which indicates that although the ability of the network to preserve knowledge is becoming weak, the ability to learn new knowledge is becoming stronger.

\subsubsection{Analysis on Different Sequences.}
In order to further evaluate the ability to prevent catastrophic forgetting on the different sequences of the proposed CGP, we conduct the experiments under the 20-task, 5-task settings on the CIFAR-100 dataset. 
The 20-Split CIFAR-100 evaluates the ability of the network to prevent catastrophic forgetting on the longer sequence.
In addition, the 5-Split CIFAR-100 evaluates the power of the network to classify more classes maintained in each task.
With the number of classes increasing, the class deviation behaves severely.
The experimental results are shown in Fig.~\ref{sequenceresult}.
It is clear that the proposed approach is significantly better than GPM in each split setting.
In particular, when we conduct the experiments on 5-Split CIFAR-100, our approach outperforms the GPM by a sizable margin at each task.
After the training, our CGP produces around 1.0\% gain in terms of the average accuracy.
The comparison results confirm that our approach has the ability to classify more different classes maintained in each task than GPM.
On 10-Split, as the experiments in the main experiment, our approach significantly outperforms the GPM.
The curve of average accuracy indicates that our method can consistently outperform GPM.
Furthermore, although GPM can achieve a comparable result to our approach on 20-Split, our method still is in general better than it.

\begin{table}[]
	\renewcommand\arraystretch{1.3}
	\renewcommand\tabcolsep{4.0pt}
	\caption{The performance of the CGP method with a various number of samples to calculate representations experimented on CIFAR-100.}
	\label{tab:number}
\begin{tabular}{ccccccc}
\hline
 \multirow{2}{*}{\textbf{Setting}}  & \multicolumn{2}{c}{20} & \multicolumn{2}{c}{125} & \multicolumn{2}{c}{200} \\ \cline{2-7} 
 & ACC & BWT & ACC & BWT & ACC & BWT \\ \hline
BR-STD & 75.03 & -1.09 & 73.43 & -2.04 & 72.81 & -2.06 \\
BR-GTL & 75.03 & -1.10 & 74.26 & -0.37 & 72.72 & -1.74 \\ \hline
\end{tabular}
\end{table}

\subsubsection{Analysis on Number of Representation Samples.}
As described in Sec.~\ref{method_simi}, we perform experiments on similarity calculation with two settings.
Our CGP selects the representation of the samples using a certain number for similarity estimation and representation matrix construction. 
We conduct experiments for BR-GTL and BR-STD including the number of 20, 125, and 200.
Tab.~\ref{tab:number} summarizes the results of this experiment.
In general, we observe that the BR-GTL has a similar performance to BR-STD in most cases, with a bit of an edge over BR-STD.
Particularly, when we just choose 20 for similarity estimation and representation matrix construction, the results are not much different in terms of average accuracy and backward transfer. 
Furthermore, a small number of samples are not enough to represent the class, which lead to inaccurately calculated bases and lead catastrophic forgetting.
When the number comes to 125, disparities are starting to appear.
The ACC and BWT of BR-STD come to 73.43\% and -2.04\%. 
Compared to the results of 20, there is a 1.6\% and 0.95\% dropping, respectively.
It is worth noting that the ACC of BR-GTL is decreasing to 74.26\%.
But there is a 0.73\% relative gain in terms of BWT, indicating that the network can preserve more knowledge.
Finally, when the number comes to 200, the performance of BR-STD and BR-GTL decreases both.
The comparison results show that directly increasing the number does not bring better performance.

\subsubsection{Analysis on $\lambda$.}
\begin{table}[]
	\renewcommand\arraystretch{1.3}
	\renewcommand\tabcolsep{4.0pt}
	\caption{Comparison of average accuracy and backward transfer on 10-Split CIFAR-100 with the $\lambda$ from 0.0 to 1.0. When $\lambda$ is 0, it represents just using classification loss.}
	\label{tab:loss}
\begin{tabular}{lllllll}
\hline
\textbf{$\lambda$} & 0.0 & 0.1 & 0.3 & 0.5 & 0.7 & 1.0 \\ \hline
ACC(\%) & 72.61 & \textbf{74.26} & 72.85 & 72.07 & 73.02 & 71.22 \\
BWT(\%) & -1.64 & \textbf{-0.37} & -1.98 & -3.16 & -2.09 & -3.91 \\ \hline
\end{tabular}
\end{table}

To evaluate the influence of  different $\lambda$ values on the final performance, we perform an ablation study with $\lambda$ from 0.0 to 1.0 under the similarity threshold of 0.7.
Table 5 shows the results of the comparison on the 10-Split CIFAR-100 dataset.
When the $\lambda$ is 0, it means that the method is just the CGP with similarity calculation, resulting in 72.61\% in terms of ACC and -1.64\% in terms of BWT.
When equipped with contrastive learning whose $\lambda$ equaling to 0.1, the ACC increases to 74.26\% and the BWT decreases to -0.37\%.
This comparison result illustrates that our CGP when equipped with contrastive learning has the stronger power to learn new knowledge and preserve the old learned knowledge.
We also note that the performance of BWT becomes worse with the increase of the $\lambda$, suggesting that as the network learns more knowledge, it gradually loses the ability to preserve the learned knowledge.

\section{Conclusion}
In this paper, we propose class gradient projection (CGP) for addressing the \textit{plasticity-stability dilemma} for continual learning. 
CGP calculates the gradient subspace from individual classes rather than tasks
Based on the CGP framework, we introduce a Base Refining (BR) algorithm as well as a contrastive loss to further alleviate catastrophic forgetting. The two components can complement each other to enhance the plasticity and stability of the model simultaneously.
The contrastive learning augments the samples to pull positive samples closer and push negative samples apart, which encourages the network to learn discriminative and robust representation.
We conduct extensive experiments on several benchmark datasets and various network architectures. The achieved results demonstrate the effectiveness of our method.

\begin{acks}
This study was supported by grants from Chinese National Science \& Technology Pillar Program (No. 2022YFC2009900/2022YFC200990\\3), the National Natural Science Foundation of China (Grant No. 62122018, No. 62020106008, No. 61772116, No. 61872064).
\end{acks}


\balance
\bibliographystyle{ACM-Reference-Format}
\bibliography{ref}
\appendix

\end{document}